# Empirical Evaluation of A New Approach to Simplifying Long Short-term Memory (LSTM)[*]

Yuzhen Lu
*Department of Biosystems and Agricultural Engineering*
*Michigan State University*
*East Lansing, Michigan 48824, USA*

*Abstract* – The standard LSTM, although it succeeds in the modeling long-range dependences, suffers from a highly complex structure that can be simplified through modifications to its gate units. This paper was to perform an empirical comparison between the standard LSTM and three new simplified variants that were obtained by eliminating input signal, bias and hidden unit signal from individual gates, on the tasks of modeling two sequence datasets. The experiments show that the three variants, with reduced parameters, can achieve comparable performance with the standard LSTM. Due attention should be paid to turning the learning rate to achieve high accuracies.

## I. Introduction

Recurrent neural networks (RNNs), as a class of deep convolutional networks (DNNs), have recently shown great promise in tackling many sequence modelling tasks in machine learning, such as automatic speech recognition [1-2], language translation [3-4], and generation of language descriptions for images [5-6]. Different from feedforward neural networks like convolutional neural network (CNN), RNNs are featured by cyclic connections with a hidden state whose activation at each time step depends on that of the previous time, which makes RNNs inherently deep along the time axis. The depth, however, makes RNNs difficult to train due to the two well-known vanishing gradient and exploding gradient problems [7-8], thus limiting their ability to learn long-term temporal dependences.

To address these problems, researchers have developed a number of techniques from the perspective of network architectures and optimization algorithms [9-11], among which the most successful one is a modified RNN architecture called Long Short-term Memory (LSTM) [9, 12]. The LSTM utilizes a memory cell that can maintain its state over time, and a gating mechanism that typically contains three non-linear gates (i.e., input, output and forget gates), to regulate the flow of information into and out of the cell, which has been proven very effective in capturing and exploiting long-range dependencies without suffering from the training hurdles that plague the conventional RNNs. Since the inception of LSTM, many improvements or alterations have been made to its structure to achieve higher performance. Gers et al. [13] added peephole connections to the LSTM that connect the memory cell to the gates so as to learn precise timing of the outputs. Sak et al. [14-15] introduced two recurrent and non-recurrent projection layers between the LSTM layer and the output layer, which resulted in significantly improved performance in a large vocabulary speck recognition task.

Adding more components in the LSTM architecture may complicate the learning process and hinder understanding of the role of individual components. Recently, researchers proposed a number of simplified variants of LSTM. Cho et al. [3] proposed a two-gate based architecture without having a separate memory cell, called Gated Recurrent Unit (GRU), in which the input and forget gates are coupled into an update gate and a reset gate is directly to applied the previous hidden state. Chung et al. [16] who made an initial comparison between LSTM and GRU, observed that the latter was comparable to or even better than the former, which, however, remains to be validated via more thorough experiments. In exploring eight simplified LSTM variants, Greff et al. [17] found that coupling the input and forget gates as in GRU and removing peephole connections did not significantly impair performance, and that the forget gate and the output activation are the critical components. These findings were corroborated by the work of Jozefowicz et al. [18] who conducted a thorough architecture search to evaluate over ten thousand different RNNs. The authors observed that the output gate was the least important compared to input and forget gates, and suggested adding a bias of 1 to the forget gate to improve the performance of LSTM. Zhou et al. [19] proposed a Minimum Gate Unit (MGU), as its name suggests, which has a minimum one gate i.e., the forget gate which is created by further coupling the update and reset gates in GRU. Through evaluations on four different sequence data, the authors found MRG with fewer parameters was on par with GRU in accuracy, but they did not perform comparisons against the standard LSTM. Very recently, Salem [20] proposed a simple approach to simplifying the standard LSTM, in which all the three gates were kept but simplified by eliminating one or two components from them, such as input signal, bias and hidden unit signal, which led to three simplified LSTM variants.

The present paper represents an effort to evaluate the effectiveness of the Salem's approach. Three simplified LSTM variants were tested and compared with the standard LSTM on the two data sets, which revealed that the simplified

---

LSTMs were capable of achieving performance comparable to the standard LSTM.

## II. LSTM Architecture

Fig. 1 show the schematic of a single LSTM memory block used in this work. The LSTM architecture is similar to that in Graves et al. [2] but without peep-hole connections. It is referred to as vanilla LSTM here and will be used for comparison with simplified variants.

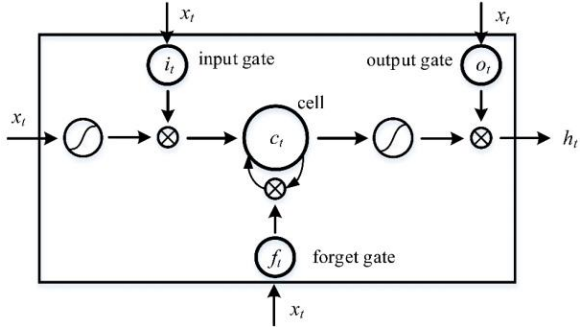

Fig. 1 A long short-term memory (LSTM) block.

The equations for the LSTM memory block are given as follows:

$$i_t = \sigma(U_i h_{t-1} + W_i x_t + b_i) \quad (1)$$
$$f_t = \sigma(U_f h_{t-1} + W_f x_t + b_f) \quad (2)$$
$$o_t = \sigma(U_o h_{t-1} + W_o x_t + b_o) \quad (3)$$
$$c_t = f_t * c_{t-1} + i_t * \tanh(U_c h_{t-1} + W_c x_t + b_c) \quad (4)$$
$$h_t = o_t * \tanh(c_t) \quad (5)$$

In these equations, $i_t$, $f_t$ and $o_t$ are the input gate, forget gate, output gate and the cell state vectors, respectively, at the current time $t$, all of which are vectors of the same size as the hidden unit signal $h$; $\sigma$ and tanh are the logistic sigmoid and hyperbolic tangent functions, respectively; $U_*$ and $W_*$ are the weight matrices, and $b_*$ are the biases; $x_t$ is the input signal at the time $t$, and the operator * denotes the element-wise vector product. The total number of parameters $N$ (i.e., the number of all the elements in $W_*$, $U_*$ and $b_*$) for the vanilla LSTM, can be calculated as follows:

$$N = 4 \times (m \times n + n^2 + n) \quad (6)$$

where $m$ and $n$ are the input dimension (i.e., number of sequences) and the number of hidden units.

A criticism of the vanilla LSTM resides in its highly complex model structure, especially the gating mechanism, which has a large number of components where computation redundancy may exist. Here, three simplifications were made to the vanilla LSTM by removing certain components from all the three gates as follows:

*1) No Input Signal*
$$i_t = \sigma(U_i h_{t-1} + b_i) \quad (7)$$
$$f_t = \sigma(U_f h_{t-1} + b_f) \quad (8)$$
$$o_t = \sigma(U_o h_{t-1} + b_o) \quad (9)$$

*2) No Input Signal and No Bias*
$$i_t = \sigma(U_i h_{t-1}) \quad (10)$$
$$f_t = \sigma(U_f h_{t-1}) \quad (11)$$
$$o_t = \sigma(U_o h_{t-1}) \quad (12)$$

*3) No Input Signal and No Hidden Unit Signal*
$$i_t = \sigma(b_i) \quad (13)$$
$$f_t = \sigma(b_f) \quad (14)$$
$$o_t = \sigma(b_o) \quad (15)$$

For simplicity of formulation, the three simplified variants above were referred to as LSTM1, LSTM2 and LSTM3. It can be seen that the three variants results in the $3mn$, $3(mn+n)$ and $3(mn+n^2)$ fewer parameters, respectively, compared to the vanilla LSTM, thus reducing the computation cost.

## III. Experimentation

The effectiveness of the three proposed variants were evaluated using two public datasets, MNIST and IMDB. Since the focus of the study was to simplify the vanilla LSTM without considerably sacrificing the performance, rather than to achieve state-of-the-art results, only the vanilla LSTM was used as a base-line model and compared with the three variants.

### A. MNIST

The dataset contains 60,000 and 10,000 images of ten-class (0-9) handwritten digits in the training and test sets, respectively, and each image has the size of 28×28 pixels. The image data were pre-processed to have zero mean and unit variance. As in the work of Zhou et al. [19], the dataset was organized in two manners or feeding LSTM networks. The first was to reshape each image as a one-dimensional vector with pixels scanned row by row, from top left corner to the bottom right corner, which resulted in a long sequence input of length 784; while the second requires no image reshaping, which treated each row of an image as a single input, thus giving an much shorter input sequence of length 28. Here, the two types of data organization were referred to as pixel-wise and row-wise sequence inputs, respectively. One could anticipate that training the pixel-wise sequence would be more time-consuming.

In the two tasks, 100 and 50 hidden units, 100 and 200 training epochs were used for the pixel-wise and row-wise sequence inputs, respectively. Other network settings were kept the same, including the batch size set to 32, RMSprop optimizer, cross-entropy loss, dynamic learning rate ($\eta$) and early stopping strategies. In particular, for the learning rate, it was set to be an exponential function of training loss $\eta = \eta_0 \times \exp(C)$ where $\eta_0$ is the learning rate coefficient, and $C$ is the training loss. For the pixel-wise sequence, two learning rate coefficients $\eta_0$=1e-3 and 1e-4 were tested for training, while for the row-wise sequence, four $\eta_0$ values of 1e-2, 1e-3, 1e-4 and 1e-5 were considered since it was much faster to train. The dynamic learning rate is directly related to instant training performance. At the initial stage, the training loss is large, thus giving a large learning rate to speed up the training process; gradually, the learning rate decreased with the loss, which helped avoid overshooting the best result. For the early stopping, the training process would be terminated if there was no improvement on the test data over consecutive 25 epochs.

## B. IMDB

The dataset consists of 50,000 movie reviews from IMDB, which are labelled into two classes according by sentiment (positive or negative), and both training and test sets contain 25,000 reviews. These reviews are encoded as a sequence of word indices based on the overall frequency in the dataset. The maximum sequence length was set to 80 among the top 20,000 most common words (longer sequences were truncated while shorter ones were zero-padded at the end). Referring to a Keras example [21], an embedding layer with the output dimension of 128, was added on top of the LSTM layer that contained 128 hidden units, and the dropout technique [22] was implemented to randomly zero 20% of embeddings in the embedding layer and 20% of rows in the weight matrices (i.e., $U$ and $W$) in the LSTM layer. The model was trained for 100 epochs. Other settings remained the same as those in MNIST data.

Training LSTMs for the two datasets were implemented by using the Keras package in conjunction with the Theano library (the implementation code and results are available at: https://github.com/jingweimo/Modified-LSTM).

## IV. RESULTS AND DISCUSSION

### A. MNIST

Table I summarizes the accuracies on the test dataset for the pixel-wise sequence. At $\eta_0$=1e-3, the *vanilla* LSTM produced the highest accuracy, while at $\eta_0$=1e-3, both LSTM1 and LSTM2 achieved accuracies slightly higher than that by the *vanilla*. The LSTM3 performed the worst in both cases.

TABLE I
THE BEST ACCURACIES OF DIFFERENT LSTM NETWORKS ON THE TEST SET AND CORRESPONDING PARAMETER SIZES OF THE LSTM LAYERS

| LSTMs | Learning rate coefficient ($\eta_0$) | | #Params |
|---|---|---|---|
| | 1e-3 | 1e-4 | |
| vanilla | 0.9857 | 0.9727 | 40,800 |
| LSTM1 | 0.9609 | 0.9799 | 40,500 |
| LSTM2 | 0.7519 | 0.9745 | 40,200 |
| LSTM3 | 0.4239 | 0.5696 | 10,500 |

Examining the training curve revealed the importance of $\eta_0$ and different responses of the LSTMs to it. As shown in Fig.2, the *vanilla* LSTM performed well in the two cases; while LSTM1 and LSTM2 performed similarly, at $\eta_0$=1e-3, which both suffered from serious fluctuations at beginning and dramatically deteriorated accuracies that leveled off at low values in the end, but decreasing $\eta_0$ to 1e-4 circumvented the problem fairly well. The LSTM3 performed differently. Both $\eta_0$=1e-3 and 1e-4 could not achieve successful training because of the fluctuation issue, suggesting that $\eta_0$ should be decreased further. As shown in Fig.3 where 200 epochs are trained, the $\eta_0$=1e-5 gave a steadily increasing pattern with the highest test accuracy of 0.7404. Despite the accuracy that was still lower than other LSTMs, it could be expected that the LSTM3 would achieve higher accuracies if long training time was allowed.

The fluctuation phenomenon observed above is a typical issue caused by a large learning rate, which is related to the different gradients on different mini-batches [23], and it can be readily resolved by turning down the learning rate but at the price of slowing training. From the obtained results, the *vanilla* LSTM seemed more resistant to fluctuations in modeling long-sequence data than the three variants, among which the LSTM3 was the most susceptible to the issue. So, a relatively small learning rat is to be used in the LSTM3, which may counteract the benefit of rapid training due to much reduced model parameters (see Table I).

Overall, these findings have showed that the three LSTM variants were capable of handling a long-range dependencies problem as the *vanilla* LSTM. Due attention should be paid in implementing them, especially LSTM3, to tuning the learning rate to achieve high accuracies.

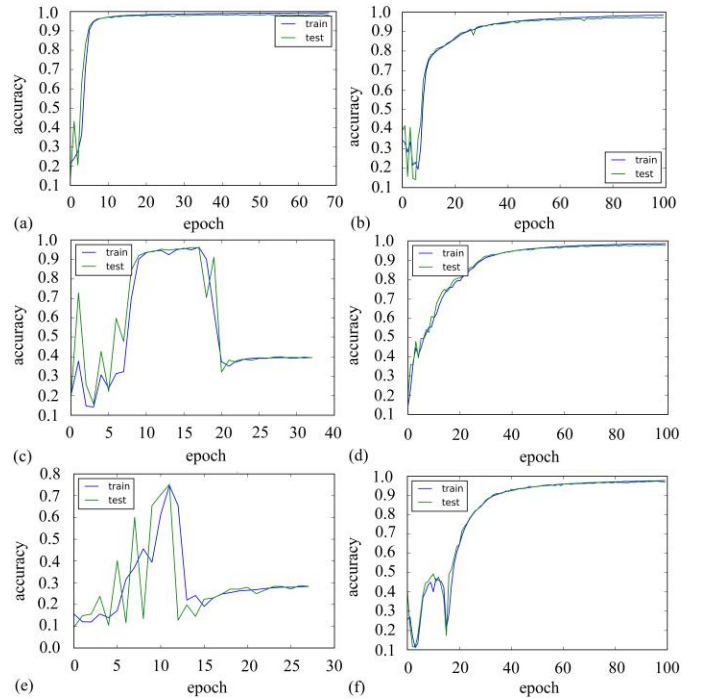

Fig. 2 Accuracies against epochs on the test data sets obtained by the vanilla LSTM (top), LSTM1 (middle) and LSTM2 (bottom), with the learning rate coefficients $\eta_0$ = 1e-3 (left) and $\eta_0$ = 1e-4 (right). The difference in epochs is due to the response to early stopping.

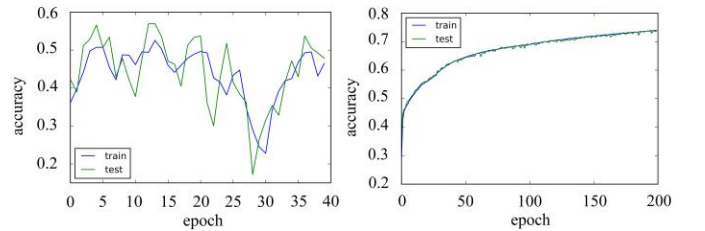

Fig. 3 Accuracies against epochs on the test dataset obtained by the LSTM3 with the learning rate coefficients $\eta_0$ = 1e-4 (left) and $\eta_0$ = 1e-5 (right). The difference in epochs is due to the response to early stopping.

Compared with the pixel-wise sequence, the row-wise sequence that has only 28 pixels in length was much easier (and also faster) to train. Table II summarizes the results. All the LSTMs achieved high accuracies at four different $\eta_0$; the *vanilla*, LSTM1 and LSTM2 performed similarly, which slightly outperformed the LSTM3. No fluctuation issues were encountered in all the cases.

TABLE II
THE BEST ACCURACIES OF DIFFERENT LSTM NETWORKS ON THE TEST SET AND CORRESPONDING PARAMETER SIZES OF THE LSTM LAYERS

| LSTMs | Learning rate coefficient ($\eta_0$) | | | | #Params |
|---|---|---|---|---|---|
| | 1e-2 | 1e-3 | 1e-4 | 1e-5 | |
| *vanilla* | 0.9506 | 0.9816 | 0.9756 | 0.9555 | 15,800 |
| *LSTM1* | 0.9820 | 0.9821 | 0.9730 | 0.9580 | 11,600 |
| *LSTM2* | 0.9828 | 0.9799 | 0.9723 | 0.9580 | 11,450 |
| *LSTM3* | 0.9691 | 0.9762 | 0.9700 | 0.9399 | 4,100 |

Among four $\eta_0$ values, the $\eta_0$=1e-3 gave the best results for all the LSTMs except LSTM2 that performed the best at $\eta_0$=1e-2. Fig. 4 shows the curves at $\eta_0$=1e-3. All the LSTMs exhibited similar training patterns, which proved the efficacy of the three LSTM variants.

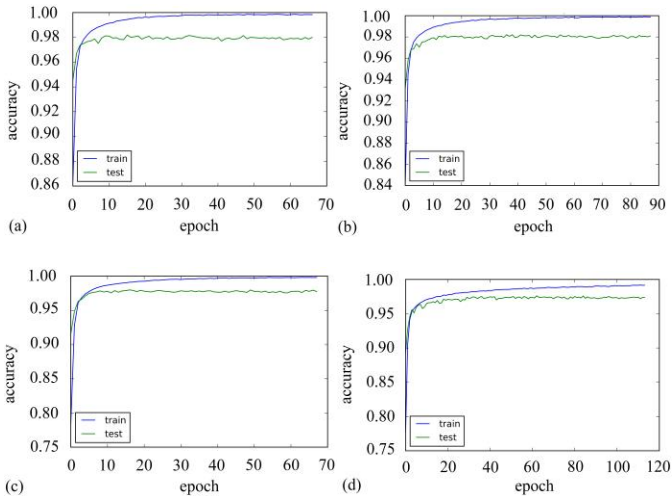

Fig. 4 Accuracies against epochs on the test dataset obtained by the vanilla LSTM (a), LSTM1 (b), LSTM2 (c) and LSTM3 (d) with the learning rate coefficient $\eta_0$ = 1e-3 . The difference in epochs is due to the response to early stopping.

Comparison of the results between the pixel-wise (long) and row-wise (short) sequence data, revealed that the three LSTM variants, especially LSTM3, performed more similarly to the *vanilla* LSTM in dealing with the short sequence data, which was probably because of the reduced complexity of the problem.

*B. IMDB*

For this dataset, the input sequence from the embedding layer to the LSTM layer was of intermediate length 128. Table III lists the training results. The *vanilla* and its three variants achieved similar accuracies, except that LSTM1 and LSTM2 showed slightly deteriorated performance at $\eta_0$=1e-2. And as in row-wise MNIST, no noticeable fluctuations were observed for the four $\eta_0$.

TABLE III
THE BEST ACCURACIES OF DIFFERENT LSTM NETWORKS ON THE TEST SET AND CORRESPONDING PARAMETER SIZES OF THE LSTM LAYERS

| LSTMs | Learning rate coefficient ($\eta_0$) | | | | #Params |
|---|---|---|---|---|---|
| | 1e-2 | 1e-3 | 1e-4 | 1e-5 | |
| *vanilla* | 0.8467 | 0.8524 | 0.8543 | 0.8552 | 131,584 |
| *LSTM1* | 0.7772 | 0.8542 | 0.8532 | 0.8550 | 82,432 |
| *LSTM2* | 0.7912 | 0.8512 | 0.8506 | 0.8510 | 82,048 |
| *LSTM3* | 0.8279 | 0.8348 | 0.8529 | 0.8548 | 33,280 |

The $\eta_0$=1e-5 consistently produced the best results for all the LSTMs, which, as shown in Fig. 5, exhibited very similar training curves.

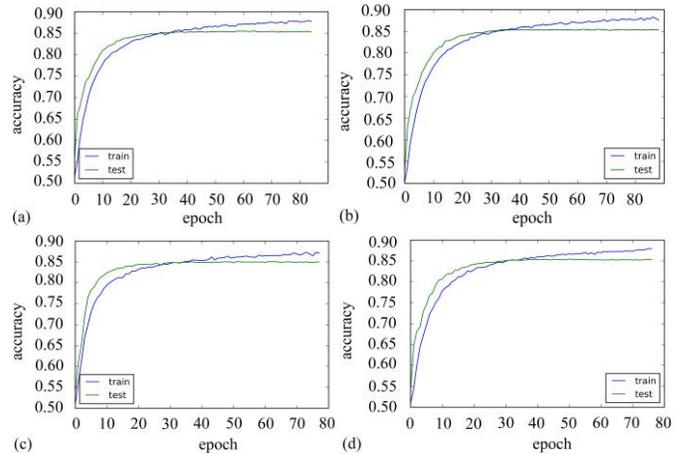

Fig. 5 Accuracies against epochs on the test dataset obtained by the vanilla LSTM (a), LSTM1 (b), LSTM2 (c) and LSTM3 (d) with the learning rate coefficient $\eta_0$ = 1e-5 .

The main benefit of the three LSTM variants is to reduce the number of parameters involved, and thus reduce the model complexity and the computation cost, which was confirmed from the statistics in the three tables above. The LSTM1 and LSTM2 had no much difference in the number of parameters since the bias contributed to a small percentage of parameters, which explained their similar performance. The LSTM3 had greatly reduced parameters since it only kept the bias, which, however, may cause problems for modeling long-sequence data. The actual reduction of parameters depends on the structure (i.e., dimension) of input sequences and the number of hidden units in the LSTM layer. This paper represents a preliminary study. Further research is needed to evaluate the three simplified variants on more extensive datasets of varied sequence length, by means of in-depth analysis of the effects of simplifications on the gating mechanism of LSTMs.

V. CONCLUSIONS

In this paper, three simplified LSTMs that were obtained by eliminating input signal, bias and hidden units from their gates in the standard LSTM, were evaluated on the tasks of modeling two sequence data of varied lengths. The results confirmed the utility of the three LSTM variants with reduced parameters, which at proper learning rates were capable of achieving the performance comparable to the standard LSTM. Further work is needed to perform more in-depth evaluation of the three variants, such as the importance of input signal, bias and hidden units in the gating mechanism.